\documentclass[fleqn,10pt]{wlscirep}
\usepackage[utf8]{inputenc}
\usepackage[T1]{fontenc}
\usepackage{siunitx}
\usepackage[normalem]{ulem}

\usepackage{capt-of}
\usepackage{caption}   
\usepackage{setspace}
\usepackage{hyperref} 

\title{Hardware-Adaptive and Superlinear-Capacity Memristor-based Associative Memory}

\author[1]{Chengping He}
\author[1]{Mingrui Jiang}
\author[1]{Keyi Shan}
\author[1]{Szu-Hao Yang}
\author[1]{Zefan Li}
\author[1]{Shengbo Wang}
\author[2]{Giacomo Pedretti}
\author[2]{Jim Ignowski}
\author[1,3,*]{Can Li}
\affil[1]{Department of Electrical and Electronic Engineering, The University of Hong Kong, Hong Kong SAR, China}
\affil[2]{Hewlett Packard Labs, Hewlett Packard Enterprise, Milpitas, CA, USA}
\affil[3]{Center for Advanced Semiconductor and Integrated Circuit, The University of Hong Kong, Hong Kong SAR, China}

\affil[*]{Email: canl@hku.hk}

\newif\ifsubmit
\submittrue

\ifsubmit
    \newcommand{\can}[1]{}
    \newcommand{\cp}[1]{}
    \newcommand{\todo}[1]{}
    \newcommand{\tocite}[1]{}
    \newcommand{\revise}[1]{}
\else
    \definecolor{comments}{rgb}{0.1, 0.66, 0.1}
    \newcommand{\can}[1]{[{\color{comments}CL: #1}]}
    \newcommand{\cp}[1]{[{\color{blue}CP: #1}]}
    \newcommand{\todo}[1]{[{\color{red}TODO: #1}]}
    \newcommand{\tocite}[1]{[{\color{red}citation-#1}]}
    \newcommand{\revise}[1]{\textcolor{orange}{#1}}
\fi

\DeclareMathOperator{\Distance}{Distance}
\DeclareMathOperator{\sgn}{sgn}

\begin{abstract}
Brain-inspired computing aims to mimic cognitive functions like associative memory, the ability to recall complete patterns from partial cues. 
Memristor technology offers promising hardware for such neuromorphic systems due to its potential for efficient in-memory analog computing. 
Hopfield Neural Networks (HNNs) are a classic model for associative memory, but implementations on conventional hardware suffer from efficiency bottlenecks, while prior memristor-based HNNs faced challenges with vulnerability to hardware defects due to offline training, limited storage capacity, and difficulty processing analog patterns. 
Here we introduce and experimentally demonstrate on integrated memristor hardware a new hardware-adaptive learning algorithm for associative memories that significantly improves defect tolerance and capacity, and naturally extends to scalable multilayer architectures capable of handling both binary and continuous patterns. 
Our approach achieves 3x effective capacity under 50\% device faults compared to state-of-the-art methods. 
Furthermore, its extension to multilayer architectures enables superlinear capacity scaling (\(\propto N^{1.49}\) for binary patterns) and effective recalling of continuous patterns (\(\propto N^{1.74}\) scaling), as compared to linear capacity scaling for previous HNNs. 
It also provides flexibility to adjust capacity by tuning hidden neurons for the same-sized patterns.
By leveraging the massive parallelism of the hardware enabled by synchronous updates, it reduces energy by 8.8× and latency by 99.7\% for 64-dimensional patterns over asynchronous schemes, with greater improvements at scale. 
This promises the development of more reliable memristor-based associative memory systems and enables new applications research due to the significantly improved capacity, efficiency, and flexibility.

\end{abstract}
\begin{document}

\flushbottom
\maketitle

\thispagestyle{empty}

\section*{Introduction}

The notion of drawing inspiration from the complex mechanisms of the biological brain has sparked a new era of breakthroughs in the field of artificial intelligence (AI)\cite{lecun2015deep}. These advancements encompass a diverse array of innovations, spanning from computational vision (CV)\cite{he2016deep} to natural language processing (NLP)\cite{vaswani2017attention}. Aside from the crucial functionality of pattern recognition, associative memory stands out as another vital aspect in the biological brain. This has been exemplified well by Pavlov's dog experiment\cite{pavlov2010conditioned}, where individuals demonstrated the ability to recall information from their memory with only partial cues. An associative memory network, also referred to as a kind of recurrent neural network, is a form of associative memory utilized for retrieving a pattern from incomplete or distorted inputs. 

\begin{figure}[htbp]
    \begin{center}
    \includegraphics[scale=0.9]{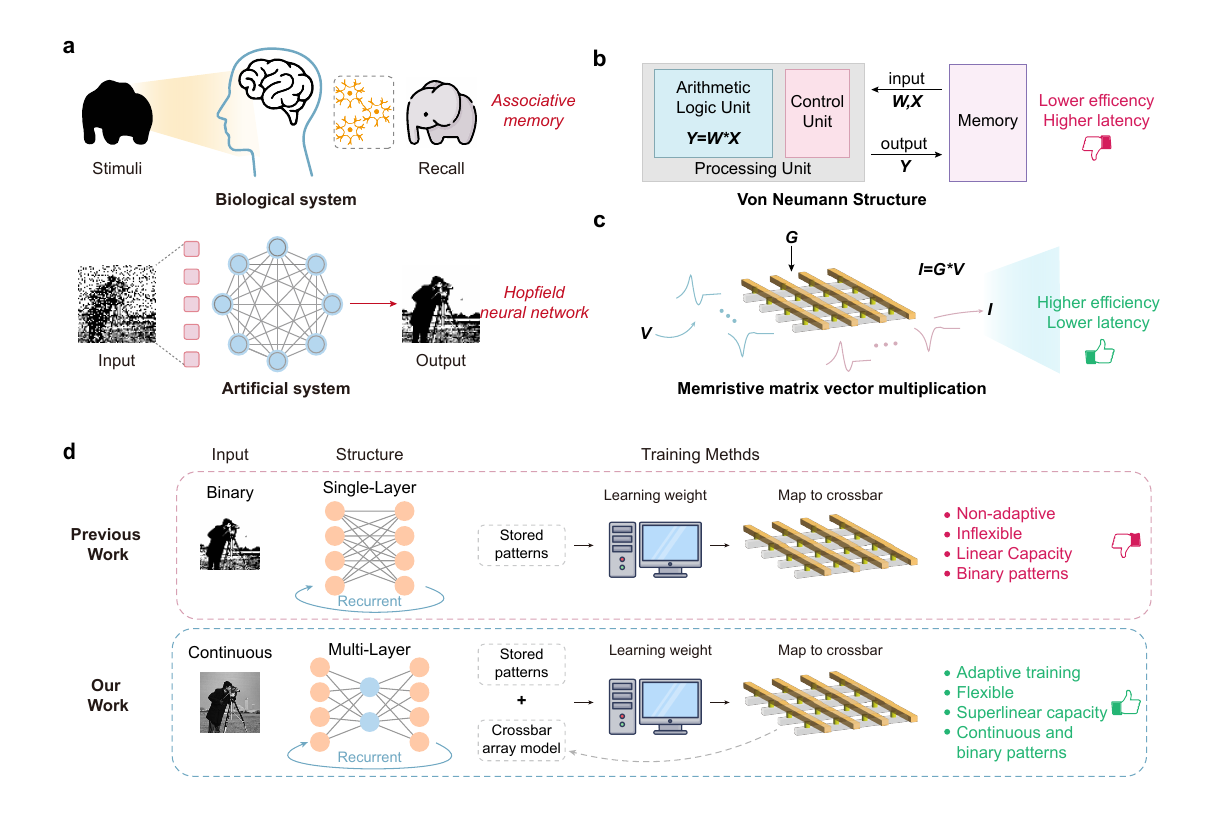}

\caption{
\textbf{Key concepts of our high-capacity memristor-based associative memory enabled by hardware-adaptive learning}: 
(a) Associative memory in the human brain and implemented by Hopfield neural networks (HNN). The human brain’s associative memory retrieves complete information from incomplete inputs. Similarly, an artificial HNN can recover stored patterns from corrupted or partial inputs. 
(b) A schematic showing the digital von-Neumann architecture computing hardware with separation of the processing unit, control unit, and memory, leading to higher latency and lower energy efficiency when implementing HNNs. 
(c) Memristor-based analog in-memory computing hardware leverages Ohm’s law and Kirchhoff’s circuit laws to perform efficient computations, achieving higher energy efficiency and lower latency for HNNs. 
(d) Compared with previous approaches that relied on offline training, single-layer structures, and binary patterns, the hardware-adaptive learning method proposed in this work enables higher capacity and defect tolerance 
, with the support of multilayer architectures and works with both binary and continuous patterns.}
	\label{Figure1}
    \end{center}
\end{figure}

Various models have been proposed in intelligent systems to implement associative memory. One of the most significant models is the Hopfield neural network (HNN)\cite{hopfield1982neural,hopfield1984neurons,hopfield1985neural,hopfield1986computing,krotov2016dense,demircigil2017model,krotov2023new}, which constitutes a fully connected network capable of utilizing local minima to store memory patterns (Figure\ref{Figure1} (a)). When a corrupted or partial pattern is inputted into the system, the system's state automatically decreases the energy based on the update rule. Consequently, the system reaches a stable state, representing a local minimum and the location where patterns are stored. As a result, the HNN can effectively retrieve the original patterns, making it a valuable associative memory system. Due to its energy-decreasing nature, the HNN finds applications in diverse fields, including solving optimization problems by introducing perturbations to reach the global minimum instead of local ones in associative memory\cite{hopfield1985neural,goto2019combinatorial}. 
However, implementing these energy-minimization processes efficiently remains challenging in classical computers due to their serial and digital processing nature and their separation of memory and process unit (Figure \ref{Figure1} (b)).

Memristors, a type of non-volatile memory, show a great potential for neuromorphic computing, including the implementation of associative memory. They offer co-located memory and computing, massive parallelism, and thus high speed and low energy consumption\cite{strukov2008missing,rao2023thousands,sharma2024linear} (Figure \ref{Figure1}c). 
Many neuromorphic computing applications have been successfully implemented using the memristor systems in the recent years, such as signal processing\cite{li2018analogue,sheridan2017sparse}, deep neural networks\cite{zhang2023edge,li2019long,wang2019situ,ambrogio2018equivalent,ambrogio2023analog,wan2022compute,zidan2018future,yao2020fully,feng2024memristor}, optimization problems\cite{jiang2023efficient,cai2020power,yang2020transiently}, scientific computing\cite{zidan2018general,le2018mixed}, and hardware security\cite{jiang2018provable,wang2024safe}. 
The Hopfield neural network can also benefit significantly from the parallel computation capability inherent in memristor systems. Many important applications of the Hopfield neural network have been successfully implemented in memristor systems, particularly in optimization tasks. Apart from the demonstration of using HNN for optimization, the implementation of HNN on memristors is inherently suited to mimic important functionalities of biological systems. 
Previous studies have demonstrated HNN implementations in various memristor technologies, including phase change memory (PCM), resistive random-access memory (RRAM) for associative memory, and ferroelectric devices in different crossbar array sizes ranging from 3$\times$3 to 128$\times$8\cite{eryilmaz2014brain,li2024emergent,wang2020memristor, yan2021ferroelectric, hu2015associative, zhou2019associative,pedretti2020spiking}.

Despite these successes, significant challenges persist (Figure \ref{Figure1}(d)). First, previous implementations are highly sensitive to device non-idealities, such as stuck-at-fault devices. This sensitivity arises because current offline
learning algorithms for associative memory in HNNs don't account for any hardware information, such as device non-idealities, which is one of reasons preventing memristive associative memory from practical use.
Second, the capacity of previous HNN models has been significantly constrained by the synaptic structure of the system. Since traditional HNNs are based on a single-layer architecture with the same number of input and output neurons, the capacity is inherently limited by the number of input neurons. For example, associative memory with Hebbian learning, the most classical learning rule, can only store up to 0.14$\times$N patterns, where N is the number of neurons\cite{hopfield1982neural}.
Therefore, once the number of input neurons is defined, the system's capacity becomes restricted, severely limiting the amount of information it can store and recall effectively.
Third, the earlier learning rule imposes stringent limitations on associating binary patterns. This restricts the functionality of associative memory, since many real-world patterns are analog values, not binary.

These limitations motivate our proposal of an online, hardware-adaptive learning algorithm that directly addresses defect tolerance, flexibility, and compatibility with analog patterns, while fully leveraging the parallelism of memristors.  By utilizing our hardware-adaptive training method, we can achieve better defect tolerance, which triples the system capacity compared to the state-of-the-art Pseudo-Inverse algorithm when 50\% of the devices are stuck on the MNIST dataset. And our approach can naturally extends to multilayer structure, significantly enhancing capacity for binary patterns and enabling associative memory for continuous patterns. For binary patterns, our method achieves a superlinear correlation between capacity and the number of neurons ($\propto N^{1.49}$) on correlated datasets like MNIST, while traditional single-layer HNNs show only a linear correlation ($\propto N^{1.06}$). Adding a hidden layer also makes the system more flexible and can reduce memristor usage by up to 95\% when storing the MNIST dataset. The multilayer structure also supports the storage of continuous patterns, achieving superlinear scalability ($\propto N^{1.74}$).  We experimentally demonstrate the successful implementation of these capabilities in our memristor-based system, which exhibits superior energy efficiency and lower latency compared to previous asynchronous update schemes. Specifically, our synchronous update can reduce processing time by 99.4\% and improve energy efficiency by a factor of 2.68× to 2.76× compared to previous asynchronous updates. Additionally, the multilayer design offers 2.53× to 3.28× higher energy efficiency and operates 1–2× faster than single-layer implementations, marking a substantial step forward in the development of scalable, energy-efficient memristor-based associative memory technologies.

\section*{Results}
\subsection*{Hopfield Neural Network For Associative Memory}

The Hopfield Neural Network was first introduced by Hopfield in 1985 \cite{hopfield1985neural}. It is characterized by being a recurrent fully connected neural network, with a decreasing energy function during the recurrent state updates.
This energy function is typically described by the Ising form energy equation:
\begin{equation}
        \label{eq1}
	H=-\frac{1}{2}\sum_{i,j}W_{ij}X_iX_j+b_iX_i
\end{equation}
Here, the $X_i$ represents the state of neuron $i$, which can be either 1 and -1, indicating whether the neuron is activated or not.  
The terms $W_{ij}$ and $b_i$ are the connection weights between different neurons and the threshold, respectively. The system updates neurons states using the following formula:

\begin{equation}
    \label{eq2}
    X_i^{n+1}=\sgn(\sum_{j=1}W_{ij}X_j^n-b_i)
\end{equation}
This update rule guarantees that the energy of the system will decrease after each update. The network will eventually reach a stable state, which is a local minimum of the energy function. 
When using the Hopfield neural network for associative memory, patterns are stored in the local minima of the system. If a corrupted pattern is input, the network will evolve towards a stable state through neuron updates, retrieving the stored pattern once stability is reached.

The patterns are stored in the local minima of the system by configuring the weights of the system. 
The most well-known learning rule for the Hopfield neural network is a biological-inspired Hebbian learning rule\cite{do1949organization}, which is a simple correlation-based learning rule, but this training method suffers from limited capacity problems. Many other learning rules, such as the Storkey learning rule \cite{storkey1999basins} and the pseudo-inverse learning rule \cite{kanter1987associative}, have been proposed to address the capacity limitations of the Hebbian learning rule. However, these methods are offline algorithms that train the weights offline based solely on stored patterns, without an adaptive training process. 
This limits their adaptability to the non-idealities when they are implemented in memristor-based systems. Due to the inherent variability and stuck-at faults in memristors, the mapped weights can significantly deviate from those learned in software. Moreover, offline training necessitates precise weight mapping, exacerbating the impact of device variations on weight accuracy.

\subsection*{Hardware-Adaptive Learning Algorithm for Associative Memory}

To address the limitations of traditional learning algorithms, we propose a new hardware adaptive learning algorithm specifically designed for memristor-based Hopfield Neural Networks. 
This method is inspired by recent research suggesting that learning rules for Hopfield networks can be conceptualized as objective function minimization\cite{tolmachev2020new}. More specifically, for states not at the energy minimum , the neuron state changes after an update. However, for states at the minimum, the neuron state remains unchanged after an update, indicating that $\mathbf{X}^{n+1}=\sgn(\mathbf{WX}^n+\mathbf{b})=\mathbf{X}^n$. Thus, at the minimum state $\mathbf{X}^n=\sgn(\mathbf{WX}^n+\mathbf{b})$(Figure S1). 
Therefore, the patterns (e.g., the $m$th pattern $\mathbf{\xi}^m$) that we stored in the system should all be the stable state of the system, accordingly, we should adjust the weight so that the system stabilizes at these stored patterns. This can be mathematically expressed as minimizing the following objective function:

\begin{equation}
    \label{eq3}
    \min_{\mathbf{W}} \sum_m\Distance\left(\mathbf{\xi}^m, \sgn(\mathbf{W}\mathbf{\xi}^m + \mathbf{b})\right)
\end{equation}
where the $\Distance$ represents the distance function used to measure the difference between the stored pattern $\mathbf{\xi}$ and the system output $\sgn(\mathbf{W\xi}+\mathbf{b})$, such as L2 distance or L1 distance. 

Since the gradient of the $\sgn$ function is difficult to compute due to its derivative being the Dirac delta function, we reformulate the problem as follows:
\begin{equation}
    \label{eq4}
    \min_{\mathbf{W}} \sum_m\Distance\left(\mathbf{\xi}^m, \tanh(\lambda(\mathbf{W\xi}^m + \mathbf{b}))\right)
\end{equation}
In this revised equation, we replace the $\sgn$ function with the $\tanh$ function to facilitate gradient descent. The parameter $\lambda$ in equation \ref{eq4} controls the steepness of the objective function; for simplicity, we set $\lambda=1$. By employing this modified loss function, we can use gradient descent to minimize the loss, thereby ensuring that the stored patterns correspond to the minima of the system.
This approach demonstrates the feasibility of training the system's weights using an objective function approach, offering a potential solution to the challenges faced by traditional learning algorithms.
 

Since the learning algorithms of the Hopfield Neural Network (HNN) can be translated into a loss function, it becomes feasible to employ gradient descent to minimize this function.
This enables the weights to be adjusted naturally to adapt to device defects when trained in-situ on hardware, or a hardware-calibrated model. 
For example, our hardware-calibrated model includes the location of stuck-at-fault devices identified in our physical memristor array\cite{mao2022experimentally}, and when training the weights with backpropagation, we apply a mask to these locations to ensure that the weights corresponding to these devices are consistently zero. Other weight values can be adjusted accordingly to compensate for the stuck-at-fault devices.
After training is complete, the optimized weights can be deployed to the physical memristor crossbar for completing associative memory tasks.

Another advantage of our adaptive learning algorithm is the ability to apply techniques from deep neural networks to the Hopfield Neural Network (HNN), such as using a multilayer structure.  
Classic HNNs are limited by their single-layer structure, where the number of neurons matches the number of input patterns, and the synaptic connections are constrained by the square of the neuron count. 
This setup heavily restricts the system's capacity and capability.
In contrast, our learning algorithm can make use of a multilayer structure, leveraging the compression capabilities of deep neural networks and the hidden layers inherent in such a structure,  and thus offers greater flexibility and the potential for improved capability and feasibility.

\begin{figure}[htbp]
	\centering   
        \includegraphics[scale=0.9]{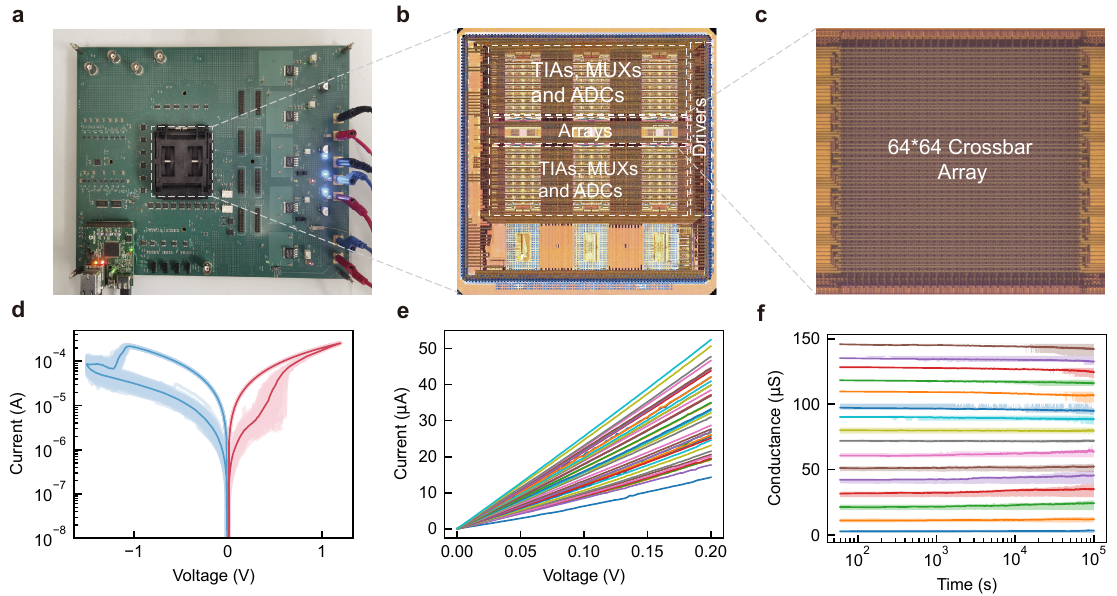}
	\caption{
    \textbf{Integrated memristor hardware for associative memory experiments :} 
    (a) Photo of our memristor-based associative memory system, comprising our integrated memristor chip, a power supply, a microcontroller, and peripheral circuitry for troubleshooting. 
    (b) The layout of our integrated memristor chip with three 64 × 64 crossbar arrays alongside fully integrated peripheral circuits—including transimpedance amplifiers (TIAs), multiplexers (MUXs), analog-to-digital converters (ADCs), and drivers. Memristor devices were fabricated via back-end-of-line (BEOL) processing in an in-house cleanroom, while CMOS peripheral circuits were fabricated using TSMC’s 180 nm technology. 
    (c) Optical microscopy image of one of the 64 × 64 memristor crossbar arrays. 
    (d) DC I-V characteristics of a one-transistor-one-memristor (1T1M) device over 100 cycles, with the dark line showing the average value. 
    (e) Linearity analysis of devices programmed to different conductance states, demonstrating excellent I-V linearity for accurate analog computing. 
    (f) Retention test for 16 conductance levels (0-\SI{150}{\micro\siemens},  \SI{10}{\micro\siemens} one step), with each level tested across 256 devices. The dark line represents the median retention, and the shaded region denotes the interquartile range, confirming stable conductance beyond $10^5$ s.}
	\label{Figure2}
\end{figure}

\subsection*{Memristor-based Associative Memory System}

We demonstrated the hardware-adaptive learning algorithm for associative memory in our integrated memristor system, as shown in Figure \ref{Figure2}(a).
The algorithm generates learned weights that represent stored patterns, which we map to the memristor crossbar by converting them to conductance values ranging from 0 to \SI{150}{\micro\siemens}.
In our experiments, we input corrupted patterns into the system via the integrated digital-to-analog converters (DACs) shown in Figure \ref{Figure2}(b).
These inputs are processed by one of our 64$\times$64 one-transistor-one-memristor (1T1M) crossbars (Figure \ref{Figure2}(c)), which performs matrix-vector multiplication in the analog domain. 
The resulting outputs are read out by integrated transimpedance amplifiers (TIAs) and analog-to-digital converters (ADCs).
The neuron states are then updated based on the sign function of this processed output.
While traditional Hopfield neural networks use asynchronous updates\cite{eryilmaz2014brain,wang2020memristor}, our memristor system implements synchronous updates (update all neurons at the same time), which better leverages the parallel computing capabilities of the memristor crossbar.

In this memristive associative memory system, the peripheral circuit and selector transistor of the integrated memristor system are designed following a 180 nm CMOS design rule, with fabrication completed by a commercial foundry.
Following CMOS fabrication, we integrated the memristor devices onto the chip in-house using back-end processes.
Full details on our integration process are available in the Method section and our previous publication\cite{sheng2019low}.
Figure \ref{Figure2}(d) illustrates the direct-current (DC) current-voltage (I-V) characteristics of the integrated memristor across 100 Set and Reset cycles, showing minimal cycle-to-cycle variation.
As demonstrated in Figure \ref{Figure2}(e), the devices can be programmed to various analog conductance states with linear I-V relationships.
These multilevel states exhibit excellent retention with no obvious systematic conductance drift, as verified by our retention test on a programmed 64$\times$64 array over $10^5$ seconds (more than one day) shown in Figure \ref{Figure2}(f).

\begin{figure}[htbp]
	\centering   
        \includegraphics[scale=0.9]
        {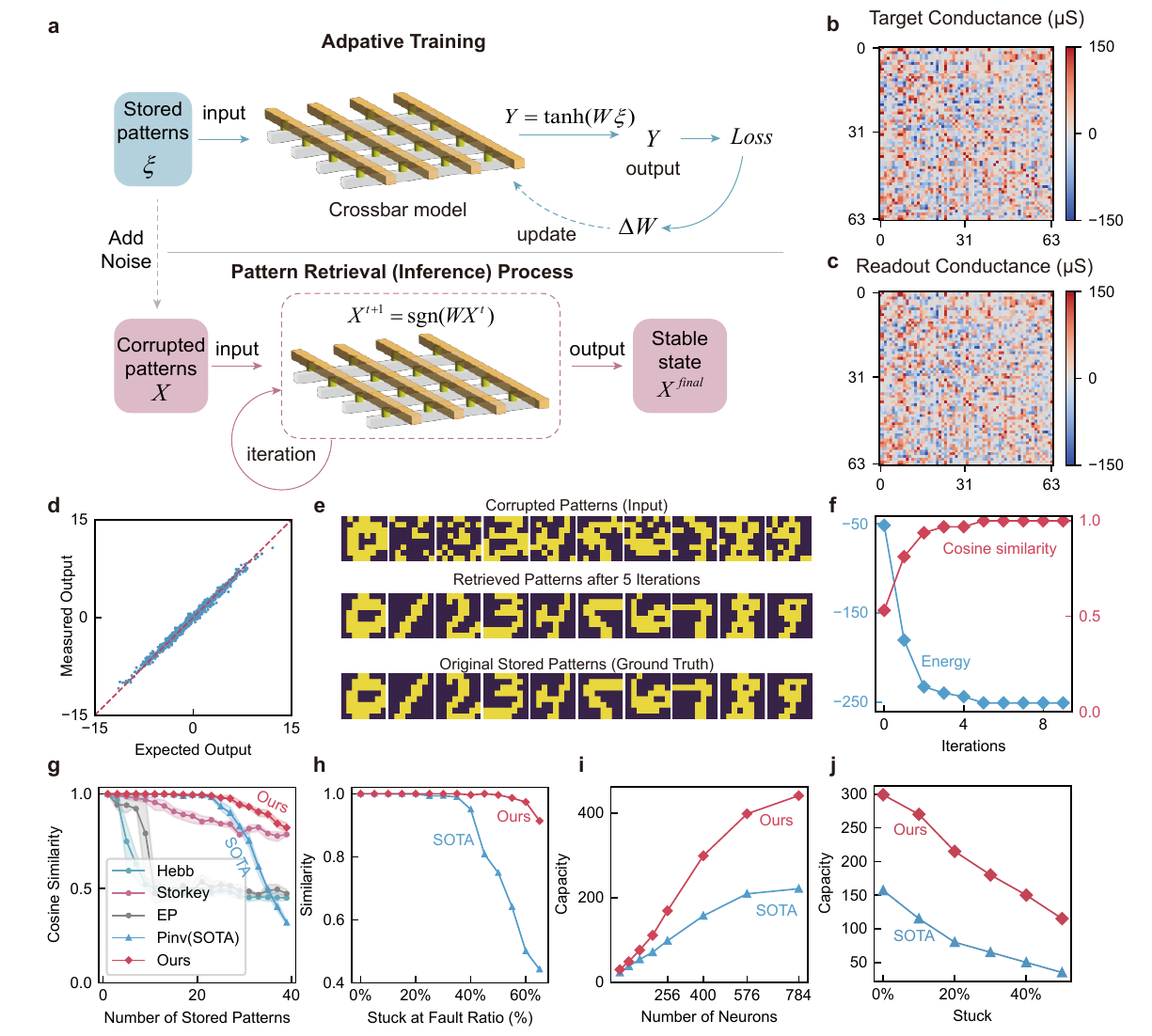}
        \caption{ 
        \textbf{Experimental demonstration of adaptive learning algorithm for single-layer HNN and performance analysis} 
        (a) Schematic illustration of the adaptive training and pattern retrieval process of our memristor-based associative memory system. A physics-based crossbar model is used to train the synaptic weights, where the device non-idealities, such as stuck-at-fault devices for that particular hardware, are taken into consideration. The inference (pattern retrieval) phase is the same as conventional HNNs, where corrupted input patterns are fed into the system, and the network state is iteratively updated until convergence to a stable state, which is then read out as the final result. 
        (b) Target conductance values representing the learned weights to be stored in the memristor array. 
        (c) Experimental readout conductance values from the memristor system after programming the target values, showing close matches with (b). 
        (d) Comparison between expected and experimental analog computing results from the crossbar, demonstrating high computing accuracy. 
        (e)  Examples of corrupted input patterns (with intentionally added noise), the corresponding stable states retrieved by the memristor system, and the originally stored patterns (ground truth). 
        (f) Energy decreases and cosine similarity increases monotonically with each iteration, eventually converging to stable equilibrium values.
        (g) Our algorithm retrieves patterns with better quality, as quantified by cosine similarity, which significantly outperforms previous methods, including the state-of-the-art (SOTA) pseudoinverse approach. 
        (h) As the fault ratio exceeds a threshold, the quality of retrieved patterns (quantized by cosine similarity) degrades dramatically, but our method exhibits superior defect tolerance compared to SOTA.
        (i) System capacity versus the number of neurons, where capacity is defined as the maximum number of patterns that can be retrieved with a cosine similarity greater than 0.99. Larger neuron counts yield higher capacity. Compared to the state-of-the-art, our method achieves approximately double the capacity. 
        (j) System capacity decreases with more stuck-at-fault devices (this experiment with 400 neurons), yet our method maintains a higher capacity, demonstrating better defect tolerance.
        }
        \label{Figure3}
\end{figure}

\subsection*{Pattern Retrieval Experiment with Hardware-Adaptive Learning Algorithm}

We perform experiments to evaluate the performance of our memristor-based associative memory system in retrieving patterns.
The stored patterns are taken from the Modified National Institute of Standards and Technology (MNIST) database, which contains 60,000 different hand-written patterns for digits 0-9. 
We randomly select different numbers of patterns and store them in the weight matrix using the previously described training algorithm, while accounting for device non-idealities in the system (Figure \ref{Figure3}a).
To fit the crossbar size, we preprocess the $28\times28$ images by cropping them to $24\times24$ and downsampling to $8\times8$ via bicubic interpolation. The images are then binarized and reshaped into $64\times1$ vectors, with each Hopfield network neuron representing one pixel.

To demonstrate how the system works, we first store ten patterns with each one taken from one of the ten digits in the MNIST dataset.
After training, each weight value is mapped to two memristor devices forming a differential pair, with their difference representing the weight value. 
Figure \ref{Figure3}(b, c) shows the target conductance map and the readout target conductance map after experimentally programming them into the crossbars, and we can see that the readout conductance is very close to the target conductance despite minor variations.
After configuring the memristor crossbar to store the patterns, we test pattern retrieval by inputting corrupted patterns into the system.
We first generate the corrupted patterns by randomly flipping 10\% of the pixels in the stored patterns, and then input these corrupted patterns into the system.
The neuron states update iteratively based on the crossbar output, with experimental values closely matching expected outputs, as shown in Figure \ref{Figure3}(d).
As the iterations progress, the system gradually reaches a stable state (neuron states do not flip anymore), which we read as the final output.
Figure \ref{Figure3}(e) shows corrupted patterns as the input and the stable state we read from the memristor system after only five iterations, which reveals no noticeable difference from the stored patterns. 
The retrieval process is further detailed in Figure S3, demonstrating how incorrect pixels are corrected after a few iterations using the synchronous update scheme. 
Most corrupted pixels are successfully recovered within two or three steps. 
We also compute the energy and cosine similarity of the neuron states to evaluate the retrieval process quantitatively.
The cosine similarity distance quantifies the difference between stored and retrieved binary patterns, with a value of one indicating identical states. Figure \ref{Figure3}(f) shows the energy and cosine similarity as functions of the number of iterations for the digit ``1''. 
As the iterations progress, the energy decreases and the cosine similarity increases, both stabilizing after five iterations. This indicates convergence to a minimum energy state and successful pattern retrieval.

One of the most important metrics for an associative memory is how many patterns can be stored, or in other words, the capacity of the system.
As the number of stored patterns increases, it becomes increasingly challenging to retrieve the correct pattern because the encoded local minima become closer to each other, leading to a higher probability of convergence to the wrong state.
Therefore, the learning algorithm has an impact on how patterns are distributed in the energy landscape, and thus on the capacity of the system.
Our experimental results demonstrated that our adaptive learning algorithm also significantly improves the system's capacity compared to previous algorithms. We randomly sample patterns from the MNIST dataset and conduct the experiment 10 times to ensure reliable comparisons.
Figure \ref{Figure3}(g) shows the cosine similarity (datapoints and solid lines showing the mean value, with shaded region representing the spread across 10 experiments\can{please confirm}\cp{yes}) between stored and retrieved patterns as a function of the number of stored patterns, where a value closer to 1 indicates better retrieval.
For the pseudo-inverse, Hebbian, and equilibrium propagation algorithms \cite{zoppo2020equilibrium, yi2023activity}, the cosine similarity between stored and retrieved patterns rapidly declines when the number of stored patterns exceeds a certain value. 
This decline leads to the "catastrophic forgetting", where all stored patterns are effectively erased once the system's capacity is surpassed \cite{mccloskey1989catastrophic, robins1998catastrophic}. In contrast, our adaptive learning algorithm demonstrates much more stable performance. 

We further evaluate the impact of stuck-at faults on the system's data integrity to demonstrate the superior defect tolerance of our adaptive learning algorithm. Figure \ref{Figure3}(h) illustrates the effect of stuck-at fault devices on the retrieval quality of the final results. For simplicity, we compare our method only with the state-of-the-art (SOTA) pseudo-inverse approach (additional comparisons with other approaches can be found in the supplementary material Figure S4-S6).  As the stuck-at-fault ratio increases, the retrieved similarity degrades; however, our algorithm maintains higher performance, tolerating fault ratios up to 50\% compared to the baseline's 35\%.

We also analyze the impact of stuck-at faults on system capacity. To evaluate the performance of different algorithms, we define system capacity as the maximum number of patterns retrievable with a cosine similarity exceeding 0.99 under a flip probability of 5\%.
Figure \ref{Figure3}(i) illustrates the scaling relationship between capacity and the number of neurons. 
While both our and the previously reported SOTA methods show a linear increase in capacity, our approach achieves twice the capacity of the state-of-the-art (SOTA) method.\can{any data about other methods in SI?}\cp{I'll add that in SI}
As the number of neurons increases, the capacity growth slows slightly due to higher pattern correlation in MNIST, which may lead to a slight performance decrease.
Figure \ref{Figure3}(j) examines how capacity degrades as the proportion of stuck-at faults increases. 
Our adaptive learning algorithm exhibits a more gradual capacity reduction compared to completing methods, effectively optimizing the use of the remaining operational devices.
Notably, at a 50\% stuck-at fault ratio, our method maintains three times the capacity of the SOTA (115 patterns versus 35),  demonstrating superior resilience against device defects.

\begin{figure}[htbp]
	\centering   
        \includegraphics[scale=0.9]{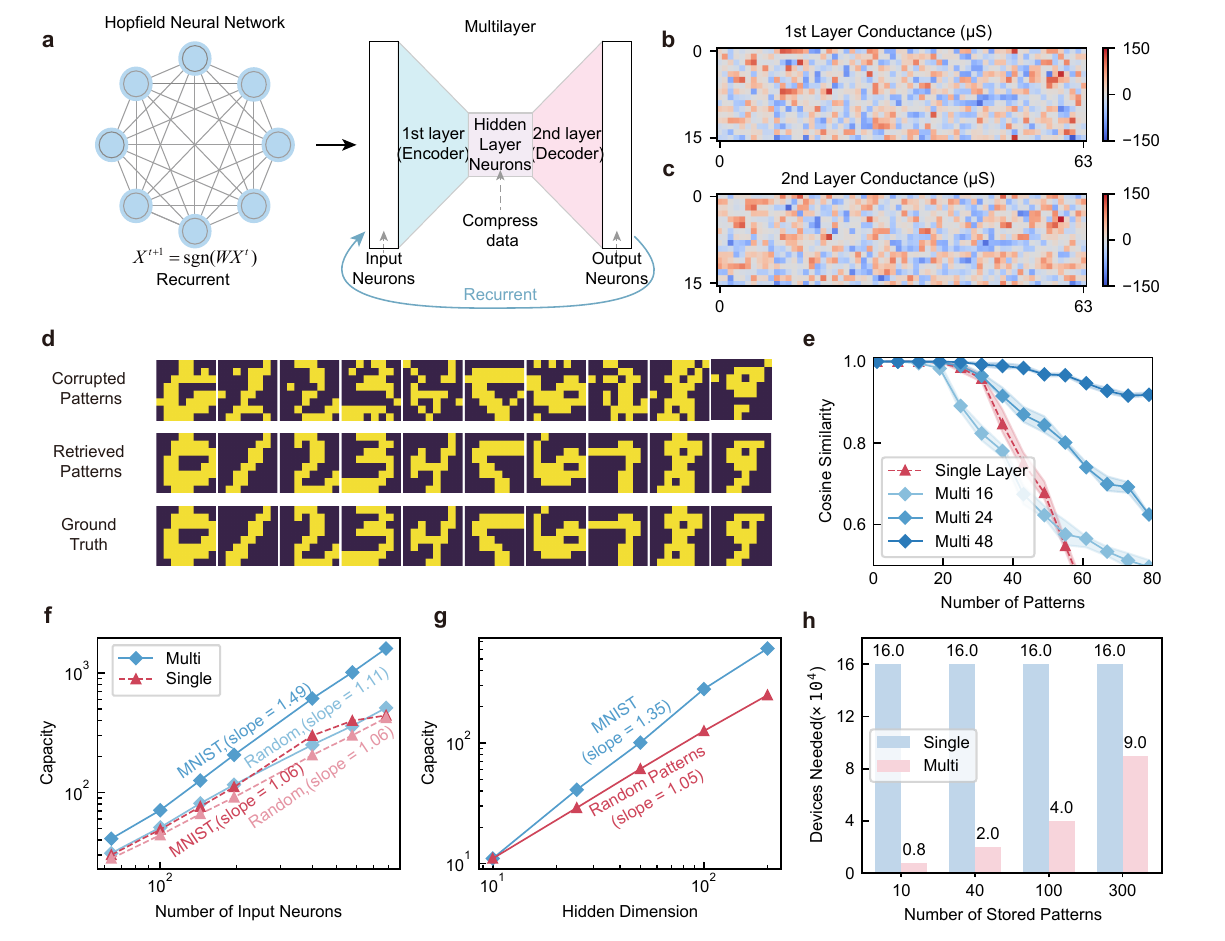}
	\caption{
    \textbf{Associative memory implemented with multilayer structure:} 
 (a) Schematic of the Hopfield Neural Network (HNN) for associative memory with a traditional one-layer fully-connected structure and the multilayer structure enabled by our adaptive learning method. 
 (b) Experimental readout conductance of the 1st layer (encoder) and (c) the 2nd (decoder) layer in a two-layer system.  
 (d) Corrupted patterns serve as the input to the associative memory (generated by flipping 10\% of the pixels in the stored patterns), the experimentally retrieved patterns, and the original stored patterns that serve as the ground truth. 
 (e) Performance comparison between the multilayer structure (with varying numbers of hidden neurons) and the single-layered HNN
 The vertical axis represents the cosine similarity between stored and retrieved patterns, while the horizontal axis indicates the number of stored patterns. 
 (f) Capacity comparison between single-layer HNN and multilayer structures (where the hidden layer dimension is half the input neuron count to make the synapse number the same as that in single-layer networks) across varying input neuron numbers. 
 (g) Capacity of the multilayer neural network as a function of hidden neuron count, with a fixed input size of 400 neurons. 
 (h) Required memristor count for storing different numbers of 20$\times$20 patterns (capacity) in single-layer and multilayer structures with 400 neurons.
 }
	\label{Figure4}
\end{figure}

\subsection*{Expanding Capacity with Multilayer Associative Memory}
While our hardware-adaptive learning algorithm significantly improves defect tolerance in single-layer Hopfield networks, conventional Hopfield Neural Networks (HNNs) remain fundamentally limited by their single-layer architecture. These traditional implementations suffer from constrained memory capacity and inflexibility due to their mandatory square weight matrix structure. 
To overcome these limitations, we propose extending the HNN framework to a multilayer architecture with recurrent iterations (Figure \ref{Figure4}(a)). This expansion is naturally supported by our loss-decreasing training algorithm and aligns with recent theoretical advances demonstrating that overparameterized neural networks can exhibit associative memory capabilities \cite{zhang2019identity, radhakrishnan2020overparameterized}. 
By incorporating hidden layers, the system gains compression capabilities, enhancing its capacity to store correlated patterns. Moreover, the hidden layers introduce adjustable capacity, allowing the system to dynamically optimize performance, as illustrated in Figure \ref{Figure4}(a).

To demonstrate the idea, we implement the multilayer associative memory using our memristor system, with each crossbar array in the chip representing one layer of the multilayer structure.
In our proof-of-concept experiment, we chose a structure with 64 input/output neurons (interfacing with 8$\times$8 patterns) and one hidden layer with 16 hidden neurons.
During the training, similar to the single-layer experiment, we trained the system to learn weights for both layers to store 10 MNIST patterns, each randomly selected from a digit. 
It is noteworthy that this configuration requires only half the synapses of a single-layered HNN. 
Figures \ref{Figure4}(b) and (c) show the experimental readout conductances for weights of the two layers of the multilayer structure, which effectively function as the encoder and decoder, respectively. 
As demonstrated in Figure \ref{Figure4}(d), the system successfully retrieves all patterns despite being presented with inputs corrupted by random 10\% pixel flips. Through iterative processing, the multilayer structure achieves perfect pattern recovery, demonstrating that our system can successfully implement multilayer structure for associative memory functionality.

To quantify the benefits of introducing a hidden layer, we conducted a comparison study between multilayer configurations with varying hidden neuron counts and traditional single-layer networks, as shown in Figure \ref{Figure4}(e).
The results show that system performance can be flexibly tuned by adjusting the number of hidden neurons, with larger hidden layers consistently yielding higher memory capacity.
Notably, the multilayer architecture achieves better performance while utilizing fewer synapses, and equivalently fewer memristors compared to conventional Hopfield Neural Networks (HNNs). 
For example, our implementation with 24 hidden neurons outperforms traditional single-layer networks while requiring only 75\% of the synapses. 

It is well established that the capacity of a single-layer HNN scales linearly with the number of neurons. 
The previous section shows that although our adaptive learning method improves the capacity of the single-layer HNN, it maintains its linear scaling relationship.
In contrast, the multilayer structure demonstrates fundamentally improved scalability due to its inherent compression capability.
For a fair comparison, we still use the number of input neurons $N$ as our reference point, and keep the number of hidden neurons to half of the input neurons, so that the number of synapses (the required  memristor devices) in the multilayer structure is equal to that of the single-layer HNN.
Figure \ref{Figure4}(f) shows the capacity comparison between our multilayer structured model and a traditional single-layer model. 
For the MNIST dataset, which contains highly correlated patterns, the capacity of the multilayer structure scales as $\propto N^{1.49}$, significantly exceeding the linear scaling observed in the single-layer HNN ( $\propto N^{1.06}$). 
Even for uncorrelated random patterns, the multilayer structure still outperforms the single-layer HNN, showing a capacity scaling of $\propto N^{1.11}$ compared to $\propto N^{1.06}$. 
The advantage of a multilayer structure is less pronounced in this case, as random patterns lack inter-pattern correlations, highlighting that the hidden layer's compression ability is critical for superlinear scalability—a capability traditional single-layer structures lack.

A significant limitation of the single-layer HNN their structural inflexibility -  once the number of input/output neurons is fixed, both the required number of synapses (memristors) and the system capacity are predetermined. 
In contrast, multilayer structures offer the flexibility to modify network architecture by adjusting the number of neurons and layers. 
We evaluated how multilayer neural network capacity scales with the number of hidden neurons ($N_h$) when the input/output neuron count remains fixed. 
As shown in Figure \ref{Figure4}(g), with 400 input/output neurons, capacity for random patterns grows linearly ($\propto N_h^{1.05}$), while for MNIST—a dataset with inherent correlations—we observe superlinear scaling ($\propto N_h^{1.35}$). 
This enhanced scaling further demonstrates the multilayer architecture's compression capability, which efficiently exploits correlations in complex data.
Since the number of synapse $N_s$ of our system is also scales linearly with hidden neurons count $\propto N_h$, which will make the capacity of the system grows linear with synapse count for random patterns ($\propto N_s^{1.05}$) and superlinear for the MNIST dataset ($\propto N_s^{1.35}$). While for the traditional single layer HNN, since the single layer structure lacks flexibility. If you want to increase the capacity, the only way you can do this is to expand the number of neurons, which will make the capacity grow only as $\propto N_s^{\frac{1}{2}}$.

Our multilayer architecture offers practical advantages for hardware implementations, especially when storing only a few high-dimensional patterns.
Traditional single-layer HNNs require a fixed number of memristors proportional to $N^2$, regardless of the number of patterns being stored. 
In contrast, our multilayer structure provides better resource efficiency. 
When storing a limited number of patterns, a compact multilayer configuration with fewer hidden neurons can be used. 
As storage demands increase, the system can be scaled by adding more hidden neurons and corresponding memristors while maintaining the same input/output neuron count (handling the same-sized patterns)- a capability unavailable in conventional HNN architectures. 
In our design, the required number of memristors is proportional to $N \times N_h$, the number of hidden neurons.  
Figure \ref{Figure4}(h) compares the required memristor device count for an associative memory that stores 20$\times$20 patterns using single-layer and multi-layer (two-layer) network structures. 
The result shows that the multilayer configuration reduces memristor requirements by 43.7\% to 95\%, depending on the number of patterns stored in memory.

\begin{figure}[htbp]
	\centering   
        \includegraphics[scale=0.9]{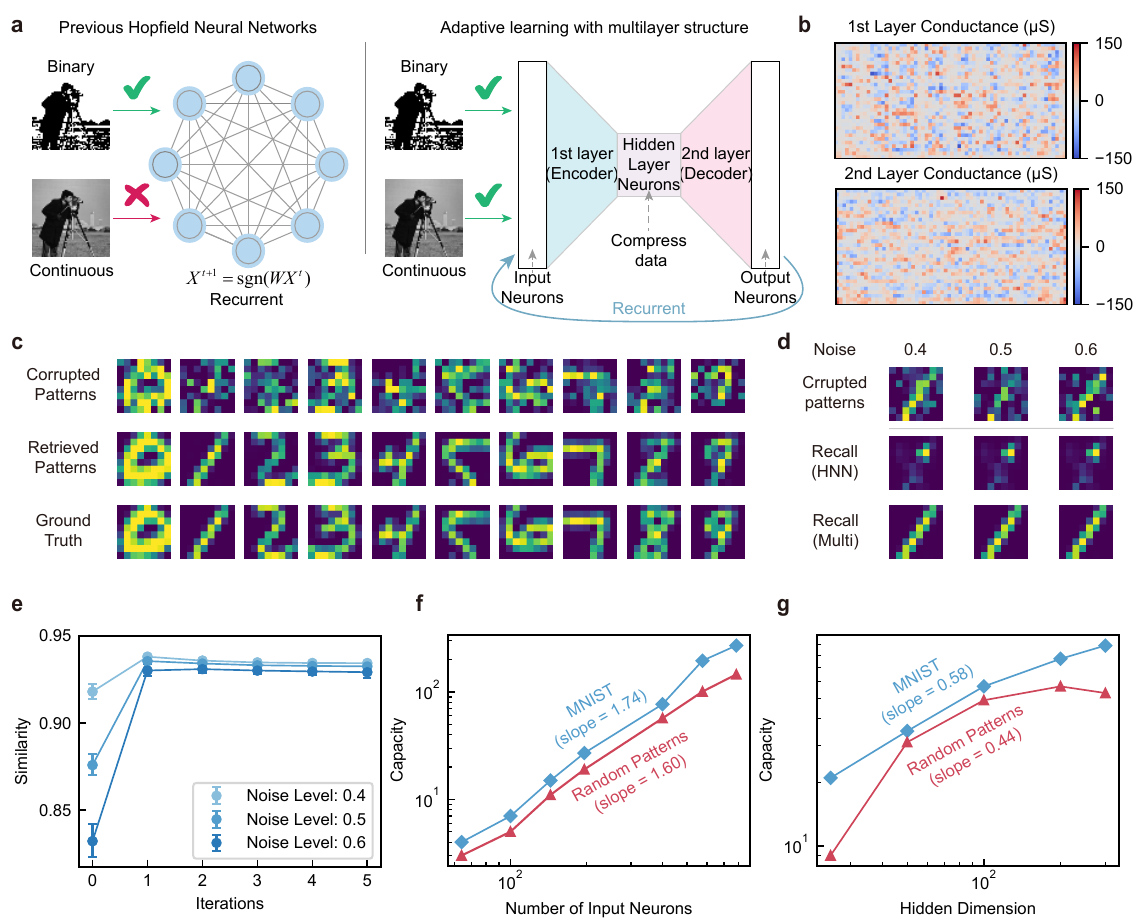}
	\caption{
    \textbf{Continuous patterns associative memory enabled by adaptive learning and multiplayer network:} 
    (a) Schematic comparison between conventional HNN that supports only binary patterns and the multilayer associative memory with adaptive learning proposed in this work that supports both binary and continuous patterns.  
    (b) Experimental readout weights (each represented by the conductance difference of two adjacent memristors) of the encoder and decoder layers in a two-layer structure. 
    (c) Hardware implementation results for associative memory with continuous patterns, where corrupted patterns are generated by adding Gaussian noise (amplitude = 0.6) to the original patterns. 
    (d) Input patterns with varying noise levels ( 0.4, 0.5, 0.6) and their corresponding retrieved patterns by single-layer HNN and multilayer structure. While the multilayer associative memory correctly retrieved the pattern, the single-layer one barely worked.
    (e) Cosine similarity between stored and retrieved patterns as a function of iteration. 
    (f) Memory capacity as a function of the number of input neurons for continuous patterns, where the number of hidden neurons is set to half the number of input/output neurons. 
    (g) Memory capacity as a function of the number of hidden neurons for continuous patterns, with a fixed total of 400 neurons.}
	\label{Figure5}
\end{figure}

\subsection*{Continuous Patterns associative Memory}

Beyond its compression capability and superior capacity, our hardware-adaptive training method addresses another critical limitation of conventional HNNs for associative memory: the inability to process continuous patterns. 
Traditional HNN learning methods with a single-layer structure are restricted to binary patterns due to their sign activation function. 
Our adaptive learning adopts a differentiable $\tanh$ function to replace the $\sgn$ function during the learning process, so it is theoretically capable of handling continuous patterns when the pattern retrieval process also uses the $\tanh$ function.
However, even with this modification, single-layer HNNs remain impractical for continuous patterns due to their extremely limited capacity(Figure S9).
Therefore, here we combine the adaptive learning and the multi-layer structure to enable the system to effectively associative continuous patterns, as illustrated in Figure \ref{Figure5}(a).

To validate this capability, we implemented the concept in our hardware prototype for continuous pattern association. 
Figures \ref{Figure5}(b) show the mapped conductance weights of both network layers. 
The retrieval process, illustrated in Figure \ref{Figure5}(c), demonstrates perfect pattern recovery even when inputs are corrupted with Gaussian noise (amplitude = 0.5). 
This confirms our architecture's effectiveness for continuous pattern processing.
Figure \ref{Figure5}(d) displays patterns with varying levels of noise alongside the corresponding retrieved patterns after iterations by both the single-layer HNN and our proposed multilayer structure. 
Despite the presence of noise, all patterns are accurately retrieved by the multilayer structure, demonstrating its robustness and superior performance in handling noisy data and ensuring reliable retrieval. In contrast, the single-layer HNN fails to retrieve the correct patterns at every noise level. Figure \ref{Figure5}(e) illustrates the retrieval similarity as a function of iteration steps. 
It is evident that, after sufficient iterations, all patterns are successfully retrieved and remain stable, confirming the effectiveness of the retrieval process of our proposed multilayer structure.

In our performance analysis with continuous patterns, we define storage capacity as the maximum number of patterns retrievable with a cosine similarity exceeding 0.99 under Gaussian noise with a standard deviation of 0.6.
Figure \ref{Figure5}(f) shows this capacity as a function of input dimensionality, with hidden neuron dimension set to half that of the original input patterns for comparison purposes.
The results demonstrate that the capacity for continuous patterns exceeds linear scaling ($\propto N$), achieving superlinear scalability of $\propto N^{1.74}$ for MNIST and $\propto N^{1.60}$ for random patterns.
Similar to the multilayer structures for binary patterns, increasing the hidden dimension enhances the system's capacity, underscoring the scalability of the multilayer structure. 
Figure \ref{Figure5}(g) demonstrates the relationship between system capacity and hidden layer dimension. The results reveal that capacity scales with hidden dimension as $\propto N^{0.58}$ for MNIST patterns and $\propto N^{0.44}$ for random patterns. This scaling behavior confirms that the system's capacity can be effectively tuned by adjusting the number of hidden neurons, providing valuable flexibility for practical implementations. The stronger scaling observed for MNIST (0.58 vs 0.44) further highlights the architecture's enhanced efficiency when processing structured, correlated data.

\subsection*{Highly Non-idealities Resilient and Efficient System}
\begin{figure}[htbp]
	\centering   
        \includegraphics[scale=0.9]{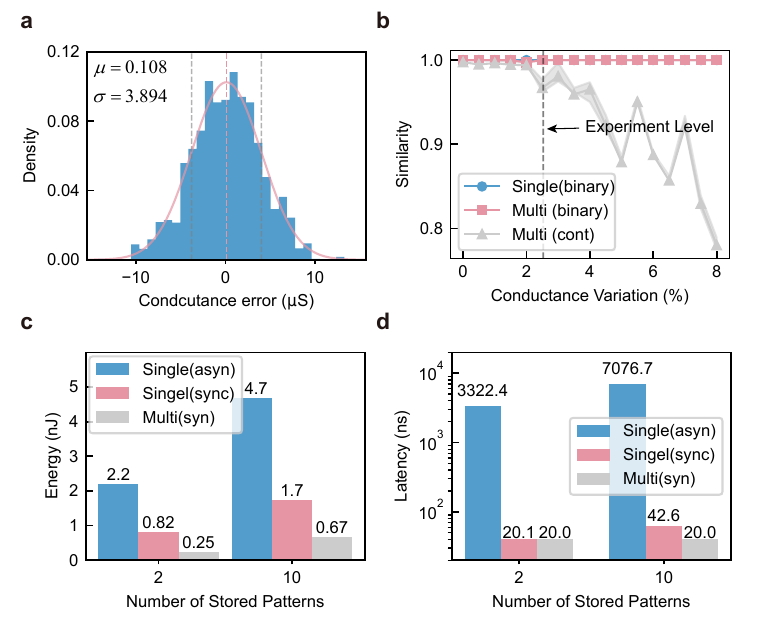}
	\caption{
    \textbf{Performance Benchmarks:} 
    (a) The typical mean and standard deviation of the programming error between the target conductance (ranging from 0 to \SI{150}{\micro\siemens}) and the actual readout conductance from our integrated memristor crossbar. 
    (b) Simulation results showing the impact of conductance variation on performance. The vertical dotted line indicates our experimental conductance variation. The experiments utilize a single-layer and multilayer structure to store 10 binary or continuous patterns, following the same setup as the previously described hardware implementation of associative memory experiments. 
    (c) Energy consumption and (d) latency of hardware implementations using different node update schemes (asynchronous updates - previous work only implemented asynchronous updates, synchronous updates), and network structure (single layer and multilayer) when storing different numbers of patterns (2 and 10). The evaluation is based on a 64-neuron system. 
    }
\label{Figure6}
\end{figure}

In analog computing, computation is affected by inevitable noise. In our system, the most significant non-idealities arise from two main sources: the mismatch between the target and mapped conductance values, and noise introduced by peripheral circuitry, such as ADC noise and line resistance. Both of these factors affect the precision of weights in the analog neural network. These noise sources manifest as deviations between the readout conductance and the intended target conductance. 
Figure \ref{Figure6}(a) illustrates the experimental distribution of these conductance differences during the mapping of a weight matrix to our hardware, with the mapped conductance range spanning 0 to \SI{150}{\micro\siemens}. The conductance difference follows a Gaussian distribution with a near-zero mean (\SI{0.108}{\micro\siemens}) and a standard deviation of \SI{3.894}{\micro\siemens}. 
Using this measured conductance variation, we simulate the effect of hardware non-idealities on experimental performance. 
Figure \ref{Figure6}(b) presents the cosine similarity as a function of conductance variation for the previously described hardware-implemented associative memory tasks. 
Our system demonstrates strong robustness to such non-idealities: for continuous-valued patterns, the similarity experiences only a minor decline under the observed experimental variations, and for binary applications, the pattern retrievals were mostly perfect because of the binarization from the sign function.

By leveraging the fully parallel update capability of memristor-based computing hardware, our method achieves significantly improved energy efficiency and reduced latency compared to conventional HNN implementations for associative memory, which can only update the node states asynchronously \cite{eryilmaz2014brain, li2024emergent, wang2020memristor, yan2021ferroelectric, hu2015associative, zhou2019associative}. 
Figure \ref{Figure6}(c) presents the energy consumption results. 
Compared to previous asynchronous implementations, our single-layer structure achieves 2.68× higher energy efficiency when storing two patterns in the memory, representing the case where only a few patterns are stored, 2.76× higher energy efficiency when 10 patterns are stored, representing more patterns are stored in the system. 
The experiments were done based on a system that stores 64-dimensional patterns, and the improvement is expected to grow with larger systems. 
Additionally, the multilayer structure demonstrates 8.8× and 7.01× improvements in energy efficiency over asynchronous single-layer associative memory for fewer (2 patterns) and more stored patterns (10 patterns).
Regarding latency, as shown in Figure \ref{Figure6}(d), our method reduces latency by 99.4\% with fewer patterns (2 patterns) for both single-layer HNN and multi-layer structure. When the storage is larger (10 patterns), we observe a 99.4\% reduction and an 99.7\% reduction over asynchronous implementations, owing to the full utilization of parallelism in analog computing.

\section*{Discussion}
In summary, we have developed a hardware-adaptive learning algorithm that successfully addresses three major limitations of previous methods: sensitivity to hardware imperfections, poor scalability and flexibility, and the restricted capability of binary patterns. We have experimentally demonstrated our approach on a memristor-based associative memory system. Our adaptive algorithm achieves significantly better capacity and defect tolerance compared to SOTA methods. On the MNIST dataset, it achieves double the capacity of existing algorithms, and triple the capacity of the SOTA baseline with a 50\% stuck-at fault ratio. By expanding the single-layer structure to a multilayer one, the system becomes more powerful and flexible. The introduction of a hidden layer enables superlinear scaling  ($\propto N^{1.49}$) for correlated patterns, significantly outperforming the linear scaling ($\propto N^{1.06}$) in single-layer networks. The hidden layer also adds flexibility and improves memristor usage efficiency (43.75\% to 95\% reduced memristor usage) compared to single-layer Hopfield Neural Networks (HNNs). The multilayer architecture not only increases capacity but also extends the functionality of associative memory systems to support continuous patterns, while also showing superlinear scalability. This makes the system more applicable to real-world scenarios, where data often exists in analog form. By leveraging the inherent parallelism of memristor arrays and using synchronous state updates, our method achieves a 2.68× to 2.76× improvement in energy efficiency and a 99.4\% reduction in processing time compared to previous asynchronous updates. These metrics are further improved due to the enhanced capabilities of the multilayer structure, achieving 2.53× to 3.28× better energy efficiency and 1–2× faster performance compared to single-layer implementations. By overcoming key limitations and harnessing the unique properties of memristors, our method demonstrates the strong potential of adaptive learning and multilayer architectures for associative memory and neuromorphic computing. Future work will explore even more diverse and complex neural network structures for associative memory and investigate their integration into broader computational frameworks and neuromorphic devices.

\section*{Methods}

\subsection*{Memristor Integration}
The memristor devices in this platform were integrated onto foundry CMOS selector and periphral circuits (with a standard 180 nm technology node) using a proprietary back-end process. The process began by removing the chip's passivation layer, followed by patterning a 2 nm layer of chromium (Cr) and a 10 nm layer of platinum (Pt) to form the bottom electrode. Next, a switching layer of 4–8 nm tantalum oxide (TaOx) was deposited via reactive sputtering. A 10 nm layer of tantalum (Ta) was then sputter-deposited as the top electrode, with an additional 10 nm platinum (Pt) layer added for protection. More details on the fabrication process can be found in Ref\cite{sheng2019low}.
\subsection*{Programming Strategy}
In this work, we employ a write-and-verify programming strategy to adjust the memristor conductance to the desired target value. Initially, we define the target conductance and set a programming error tolerance range of \SI{5}{\micro\siemens}. The target conductance is derived from the original matrix and scaled to the desired conductance range (\SI{0}{\micro\siemens} to \SI{150}{\micro\siemens}) without quantization. During the write-and-verify process, the conductance of each device in the memristor array is first measured using a 0.2 V read voltage. If the measured conductance falls within the defined tolerance range of the target conductance, no further action is taken for that memristor. However, if the conductance deviates from the target by more than the allowable tolerance, a RESET or SET voltage is applied to decrease or increase the conductance, respectively, bringing it closer to the target value. Throughout the programming iterations, both the RESET and SET voltages, along with the gate voltage, are gradually increased.

\subsection*{Hardware-Adaptive Training Algorithm}
For the single-layer structure, the L2 distance is used to measure the discrepancy between the retrieved patterns and the original patterns in the system. To simplify the calculations, we set the parameter $\lambda=1$ , which affects only the steepness of the objective function. Consequently, the final loss function is defined as:
\begin{equation}
    \label{eq5}
    Loss=\frac{1}{m}\sum_m\left(\mathbf{\xi}^m-\tanh(\mathbf{W\xi}^m + \mathbf{b})\right)^2
\end{equation}
During training, we set the learning rate to  $3\times 10^{-2}$,  and run the process for a maximum of 10,000 steps. Training terminates early if the loss drops below $1\times 10^{-8}$.

For the multi-layer structure, the training loss function is modified to:
\begin{equation}
    \label{eq6}
    Loss=\frac{1}{m}\sum_m\left(\mathbf{\xi}^m-\mathbf{f}(\mathbf{\xi}^m)\right)^2
\end{equation}
Here, $\mathbf{f}(\xi^m)$ represents the output of the multi-layer network. The learning rate is set to $3\times10^{-4}$, with a maximum of 60,000 epochs, stopping when the training loss falls below $1e^{-8}$, we will stop the training. 
During training, we use Root Mean Square Propagation (RMSprop) as the optimizer to reduce the loss. 
To handle stuck-at-fault devices, we make the weights at the corresponding locations untrainable and set their values to zero during training.

\subsection*{PCB-based interface board with memristor-chip}

A custom-designed printed circuit board (PCB) supports both programming and pattern retrieval operations on the chip. 
The PCB serves as an interface between the chip and external off-the-shelf components. 
This interface board includes a microcontroller that enables communication with a computer, as well as a power supply that delivers the necessary operating voltage via power supply or on-board digital-to-analog converters (DACs). 
Most peripheral circuits, such as transimpedance amplifiers and analog-to-digital converters (ADCs), are integrated on-chip, while the board mainly provides probes for troubleshooting purposes. 
The chip itself is wire-bonded to a chip-holder on the testing board.

\section*{Data availability}
The data supporting the findings of this study are included in the main text and the Supplementary Information. Additional data related to this study are available from the corresponding author upon reasonable request.
\section*{Code availability}
The custom code developed for this study is openly available in the following GitHub repository: 

\href{https://github.com/hecp2025/MultilayerAssociateMemory}{https://github.com/hecp2025/MultilayerAssocaiteMemory}

\bibliography{references}

\section*{Acknowledgements}
This work was supported in part by Research Grant Council of Hong Kong SAR (C7003-24Y, 27210321, C1009-22GF, T45-701/22-R), National Natural Science Foundation of China (62122005), ACCESS – an InnoHK center by ITC, and Croucher Foundation. 
\section*{Author contributions}
C.H. and C.L. conceived the initial idea for this study. C.H., M.J., K.S., S.Y., and Z.L. conducted the experiments. C.H. performed simulations, data analysis, and comparative evaluations.S.W. help draw the figures. C.L. supervised the project. The manuscript was written by C.H., G.P., J.I., and C.L., with contributions and feedback from all authors.

\end{document}